\documentclass[twocolumn]{cinc}
\usepackage{graphicx}
\usepackage{booktabs}
\usepackage{amsmath,amssymb}
\usepackage[hidelinks]{hyperref} 
\usepackage{siunitx}

\begin{document}
\bibliographystyle{cinc}

\title{On the role of the tokenizer in ECG transformer models}


\author {Jiawei Li$^{1}$, Fabio Bonassi$^{1}$, Johan Sundström$^{1}$, \\Thomas B. Schön$^{1}$, Antonio H. Ribeiro$^{1,2}$\\
\ \\ 
 $^1$ Uppsala University, Uppsala, Sweden \\
$^2$  Scilifelab, Uppsala, Sweden}

\maketitle

\begin{abstract}
Tokenization determines both the physiological content presented to an ECG Transformer and the sequence over which attention operates. We compare eight tokenization strategies across Transformer, Informer, Reformer, and FEDformer on the nine-label CPSC2018 classification task. The input projection and principal backbone capacity are controlled to isolate the effect of token construction. Median-beat and HeartLang tokenization achieve mean macro-AUCs of 0.893 and 0.889 across the four backbones, compared with 0.822 and 0.824 for point-wise and patch-wise tokenization. Pooling the two physiology-aware representations yields an 8.2\% relative improvement in macro-AUC. They also reduce mean sequence length from 1,250 to 158 tokens and mean peak training memory from 5.21 to 0.27~GB.  The results show that aligning tokens with ECG morphology can improve both predictive performance and memory efficiency without increasing backbone capacity. The source code is available on \url{https://github.com/LeeJarvis996/ecg_tokenizer}.
\end{abstract}

\section{Introduction}

Transformer architectures have achieved strong performance in general time-series modelling and are increasingly being adopted for electrocardiogram (ECG) analysis. Nevertheless, their performance on ECG classification often remains suboptimal, usually falling behind state-of-the-art convolutional neural networks and state-space models~\cite{al2026benchmarking}. While previous work has primarily focused on improving attention mechanisms~\cite{guoqi2026decentralized} or scaling model size~\cite{gu2026cardiac}, we argue that tokenization is also an important design choice. An ECG is a quasi-periodic, multi-lead signal whose diagnostic information is encoded in the morphology of the P--QRS--T complex, beat-to-beat rhythm, and spatial relationships among leads. Point-wise tokenization preserves the original sampling resolution but produces long sequences, whereas fixed-length patches may divide physiologically coherent waveforms across token boundaries. Token count also directly affects computational cost, with full self-attention scaling quadratically with sequence length. 

In this study, we examine the role of tokenization in ECG Transformer models by evaluating eight strategies across four representative backbones, Transformer~\cite{3295222.3295349}, Informer~\cite{haoyietal-informer-2021}, Reformer~\cite{kitaev2020reformer}, and FEDformer~\cite{zhou2022fedformer}, on the CPSC2018 classification task. The evaluated representations span point-wise, lead-wise, fixed-length patch, stochastic segment, BIOT-inspired~\cite{biot} segment, ECG-Byte~\cite{han2024ecgbytetokenizerendtoendgenerative}, HeartLang heartbeat~\cite{jin2025reading}, and median-beat tokenization, while the input projection is unified and the principal backbone hyperparameters are held constant across experiments. This design allows us to investigate how token definition and sequence length influence predictive performance and computational efficiency. 

Averaged across the four backbones, median-beat and HeartLang tokenization achieve macro-AUCs of 0.893 and 0.889, respectively. Pooling these two physiology-aware tokenizers gives a mean macro-AUC of 0.891. Conventional point-wise and patch-wise tokenization achieve corresponding means of 0.822 and 0.824, with a pooled mean of 0.823. The pooled improvement is 8.2\% relative. Compared with the pooled point-wise and patch-wise group, the physiology-aware group reduces mean sequence length from 1,250 to 158 tokens and mean peak training VRAM from 5.21 to 0.27~GB, corresponding to reductions of 87.4\% and 94.8\%, respectively. 
Although our evaluation is limited to a single dataset, the results suggest that preserving physiologically meaningful structure is crucial and that a simple tokenizer choice can yield substantial classification performance and efficiency gains.

\section{Methods}

\subsection{Tokenization Definition}

For one ECG record, let $\mathbf{X}\in\mathbb{R}^{\text{T}\times \text{C}}$ denote $T$ samples from $C$ leads. A tokenizer $\tau$ transforms the signal into an ordered sequence of \emph{token units},
\begin{equation}
\mathcal{U}_{\tau}(\mathbf{X})
=(\mathbf{u}_{\text{1}},\ldots,\mathbf{u}_{\text{L}}),\qquad
\mathbf{u}_{\text{k}}\in\mathbb{R}^{\text{P}},
\end{equation}
where $\mathbf{u}_{\text{k}}$ is the content of the $k$-th token unit, $L$ is the sequence length, and $P$ is its tokenizer-specific dimension. 
The construction of $\mathbf{u}_{\text{k}}$ determines its physiological meaning, while the ordered sequence of $L$ units defines the axis along which the Transformer models token-to-token relationships.

Each token unit is converted to the common model dimension $D$ by an input projection $\phi_{\tau}$, and optional lead and positional information is then added. We define the resulting \emph{token embedding}, i.e., the vector actually passed to the Transformer, as
\begin{equation}
\mathbf{e}_{\text{k}}
=\phi_{\tau}(\mathbf{u}_{\text{k}})
+\mathbf{e}_{\mathrm{lead}}(\text{k})
+\mathbf{e}_{\mathrm{temp}}(\text{k})
\in\mathbb{R}^{\text{D}},
\end{equation}
where $\mathbf{e}_{\mathrm{lead}}(\text{k})$ and $\mathbf{e}_{\mathrm{temp}}(\text{k})$ denote the lead embedding and the temporal embedding associated with the $k$-th token unit, respectively. 
The embedding sequence $\mathbf{E}=(\mathbf{e}_{\text{1}},\ldots,\mathbf{e}_{\text{L}})\in\mathbb{R}^{\text{L}\times \text{D}}$ is then processed by a Transformer.

\begin{table}[t]
\caption{Token units for a \qty{10}{\second} , \qty{100}{\hertz}, 12-lead ECG.}
\label{tab:token_units}
\centering
\scriptsize
\setlength{\tabcolsep}{10pt}
\begin{tabular}{@{}lccp{3.45cm}@{}}
\toprule[1pt]
Tokenizer & $L$ & $P$ & Physiological meaning of one unit \\
\midrule[0.6pt]
Point-wise & 1000 & 12 & Simultaneous sample across all leads \\
Lead-wise & 12 & 1000 & Complete waveform from one lead \\
Patching & 1500 & 16 & Overlapping \qty{0.16}{\second} segment from one lead \\
Stochastic & 192 & 128 & Resampled \qty{0.8}{\second}--\qty{1.6}{\second} segment from one lead \\
BIOT-inspired & 228 & 100 & Fixed \qty{1}{\second} segment from one lead \\
ECG-Byte & 1020 & ID & BPE unit over quantized ECG symbols \\
HeartLang & 256 & 96 & One heartbeat from one lead \\
Median beat & 60 & 12 & One relative sample of a representative multi-lead beat \\
\bottomrule[1pt]
\end{tabular}
\end{table}

\subsection{Tokenizer Architectures}

Let us now briefly introduce the eight tokenizer baselines that we will compare. All dimensions below refer to our \qty{100}{\hertz}, \qty{10}{\second}, 12-lead ECG input ($T=1000$, $C=12$). Here, $k$ follows the token-unit index introduced in Section~2.1; for lead-specific tokenizers, the lead index $c$ and within-lead index $n$ are flattened into $k$ in lead-major order.

\textbf{Point-wise.} Each simultaneous multi-lead sample is one token unit,
$\mathbf{u}_{\text{k}}=\mathbf{X}_{\text{k,:}}$ for $k=1,\ldots,T$. Hence, $L=T$ and $P=C$, and attention models temporal relations at the original sampling resolution. Since each unit mixes all leads, it has no unique lead identity.

\textbf{Lead-wise.} Inspired by iTransformer, the input axes are inverted so that one complete lead is one token unit,
$\mathbf{u}_{\text{k}}=\mathbf{X}_{\text{:,k}}$ for $k=1,\ldots,C$. Hence, $L=C$ and $P=T$, and attention operates across leads rather than time points.

\textbf{Patching.} Each lead is divided into overlapping windows of length $p$ with stride $s$, with one stride replicated at the right boundary. Let $N_p=\lfloor(T+s-p)/s\rfloor+1$. The unit indexed by $k=(c-1)N_p+n$ is the $n$-th patch from lead $c$, giving $L=CN_p$. We choose $p=16$ and $s=8$ via grid search.

\textbf{Stochastic segments.} For each lead, up to $K$ intervals $[a_n,b_n]$ are sampled, where the segment length satisfies $b_n-a_n\in[\ell_{\min},\ell_{\max}]$ and the start-point increment satisfies $a_{n+1}-a_n\in[s_{\min},s_{\max}]$. The unit $\mathbf{u}_{\text{k}}$, indexed by $k=(c-1)K+n$, is obtained by linearly resampling $\mathbf{X}_{\text{a}_\text{n}:\text{b}_\text{n},\text{c}}$ to $P$ samples. A seeded template fixes the same boundaries across leads and throughout a run. The sequence contains at most $CK$ units and is padded to $L_{\max}=CK$ for batching. We choose $K=16$, $P=128$, $[\ell_{\min},\ell_{\max}]=[80,160]$, and $[s_{\min},s_{\max}]=[50,70]$ via grid search.

\textbf{BIOT-inspired segments.} Following~\cite{biot}, each lead is divided into fixed windows of length $p=100$ with stride $s=50$. This gives $N_b=\lfloor(T-p)/s\rfloor+1=19$ windows per lead. The index $k=(c-1)N_b+n$ denotes the $n$-th window from lead $c$, yielding $L=CN_b=228$.

\textbf{ECG-Byte.} The central 2s interval is resampled from 100 to \qty{250}{\hertz}, normalized using training-set percentiles, and quantized into 26 amplitude symbols. The lead-major symbol stream is compressed using the pretrained 3,500-entry byte-pair encoding vocabulary~\cite{han2024ecgbytetokenizerendtoendgenerative}. Each $\mathbf{u}_{\text{k}}$ is a discrete vocabulary index. The resulting sequence is capped at $L_{\max}=1020$ and padded when shorter.

\textbf{HeartLang.} Following~\cite{jin2025reading}, QRS complexes are detected once on the lead II, and the resulting beat boundaries are shared across all leads. Let $\mathbf{b}_{\text{c,n}}\in\mathbb{R}^{\text{96}}$ denote the $n$-th heartbeat from lead $c$ after cropping or zero-padding. These lead--beat units are ordered lead-major such that $\mathbf{u}_{k}=\mathbf{b}_{\text{c,n}}$  and the variable-length sequence is capped and padded to $L_{\max}=256$.

\begin{figure*}[!t]
\centering
\includegraphics[width=\textwidth]{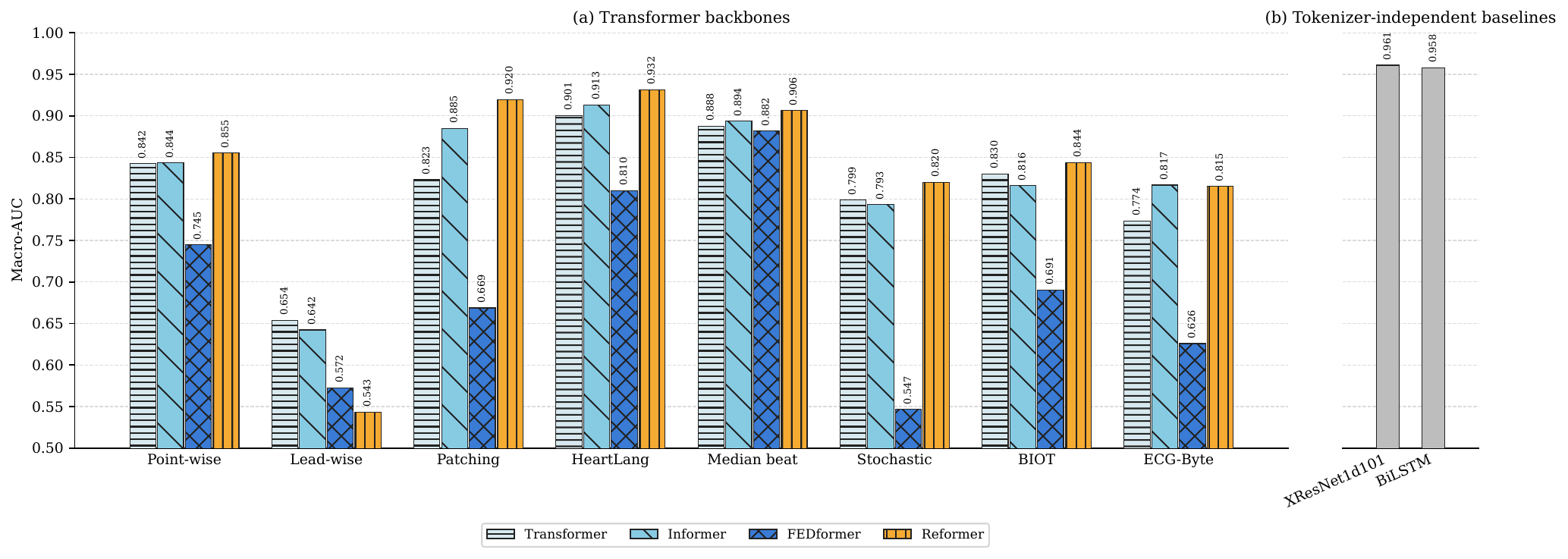}
\caption{CPSC2018 test macro-AUC under eight tokenizer configurations. The tokenizer-independent convolutional and recurrent baselines are shown separately.}
\label{fig:main_results}
\end{figure*}

\textbf{Median beat.}
R peaks are detected on lead II, and R-centred multi-lead beats of length $W$ are extracted. Their element-wise median yields a representative heartbeat
$\widetilde{\mathbf{X}}\in\mathbb{R}^{\text{L}\times \text{C}}$.
Each relative time point constitutes one token unit,
$\mathbf{u}_{\text{k}}=\widetilde{\mathbf{X}}_{\text{k,:}}$ for
$k=1,\ldots,L$.
Attention therefore models morphological relationships within a representative heartbeat rather than relationships among individual beats. As in point-wise tokenization, each unit contains measurements from all leads and consequently has no unique lead identity. The beat length was set to $L=60$ samples based on a grid search, and we found that longer windows did not improve performance.

\section{Results}
\subsection{Experiment Setup}

\noindent\textbf{Baselines.} We evaluate four \textit{Transformer backbones}: Transformer~\cite{3295222.3295349}, Informer~\cite{haoyietal-informer-2021}, Reformer~\cite{kitaev2020reformer}, and FEDformer~\cite{zhou2022fedformer}.
For \textit{tokenizer-independent} context, we also consider the convolutional XResNet1d101 and recurrent bidirectional LSTM baselines. 

\noindent\textbf{CPSC2018 dataset.}  We downsample each recording to 100~Hz and retain or resample it to 10~s. The task predicts nine diagnostic labels: AFIB, VPC, NORM, 1AVB, CRBBB, STE, PAC, CLBBB, and STD. The dataset was divided into training, validation, and test sets in a patient-stratified 8:1:1 ratio. A global standardizer is fitted on the training set and then applied to all splits.

\noindent\textbf{Implementations.} For a controlled comparison, every Transformer backbone uses embedding dimension $D=128$, three encoder layers, and eight attention heads. Continuous token units are mapped from $P$ to $D$ by a linear input projection, whereas ECG-Byte uses a learned vocabulary embedding. Following HeartLang, lead-specific tokens receive a learned embedding indexed by lead identity. The positional term combines a learned absolute token-position vector with a learned embedding of the temporal index.  Models are optimized with RAdam and binary cross-entropy at a learning rate of $10^{-3}$ for at most 300 epochs, with early stopping after 15 epochs without validation macro-AUC improvement. All experiments are conducted on a single NVIDIA A40 GPU.

\subsection{Main Results}

Figure~\ref{fig:main_results} reports the best test macro-AUC for each available backbone--tokenizer pair. The CNN and recurrent baselines reach 0.958--0.961 macro-AUC and outperform all evaluated Transformer combinations. Among the latter, the strongest combination is Reformer with HeartLang, reaching 0.932 macro-AUC and 0.607 macro-F1. Reformer with patching ranks second in macro-AUC (0.920), while Informer with HeartLang obtains 0.913 macro-AUC and the highest macro-F1. HeartLang is the best tokenizer for Transformer, Reformer, and Informer.

Median-beat tokenization is the most consistent compact representation. It is second best for Transformer and Informer, remains competitive for Reformer, and is the strongest FEDformer input by a large margin, improving its macro-AUC from 0.745 with point-wise tokens to 0.882 with only 60 tokens. 
Patching is highly backbone-dependent, which performs  poorly with FEDformer. 
ECG-Byte is less competitive, which may reflect both its restriction to the central 2~s and the observed truncation at its 1,020-token cap. 
Lead-wise inversion performs worst or near-worst across all four backbones, contrasting with its strong performance in general time-series forecasting.

\begin{table}[b]
		\caption{Complete model efficiency results. Each entry reports end-to-end inference latency in ms/ECG followed by peak allocated training VRAM in GB (latency/VRAM). The lowest value is bold and the second lowest is underlined. Latency is averaged over ten repetitions; VRAM is measured for a complete training step.}
	\label{tab:model_efficiency}
	\centering
	\footnotesize
		\setlength{\tabcolsep}{5pt}
	\begin{tabular}{@{}lrrrr@{}}
		\toprule[1pt]
		Tokenizer & Transformer & Reformer & Informer & FEDformer \\
		\midrule[0.6pt]
		Point-wise & 1.35/8.52 & 1.95/3.41 & 0.57/1.08 & 0.86/0.60 \\
			Lead-wise & \textbf{0.06}/\textbf{0.03} & \textbf{0.15}/\textbf{0.06} & \textbf{0.10}/\textbf{0.03} & \textbf{0.21}/\textbf{0.04} \\
		Patching & 3.11/18.86 & 3.78/6.59 & 0.89/1.75 & 0.90/0.87 \\
		Stochastic & 4.17/0.40 & 4.26/0.45 & 4.26/0.21 & 4.90/0.18 \\
			BIOT-inspired & \underline{0.12}/0.54 & \underline{0.29}/0.54 & \underline{0.16}/0.25 & \underline{0.84}/0.20 \\
		ECG-Byte & 11.79/8.87 & 12.18/3.56 & 11.01/1.10 & 11.31/0.61 \\
		HeartLang & 4.59/0.66 & 4.67/0.59 & 4.62/0.27 & 5.22/0.22 \\
			Median beat & 3.87/\underline{0.08} & 3.96/\underline{0.16} & 3.95/\underline{0.08} & 4.64/\underline{0.10} \\
		\bottomrule[1pt]
	\end{tabular}
\end{table}

\subsection{Analysing efficiency}

Tables~\ref{tab:model_efficiency} and~\ref{tab:tokenizer_efficiency} separate complete model cost from tokenizer-only latency. Measurements use an NVIDIA A40, batch size 32, three warm-up iterations, and ten timed repetitions. 
Patching produces 1,500 tokens and requires 18.86~GB with Transformer, whereas lead-wise tokenization requires at most 0.06~GB across the four backbones. 
The compact physiological representations also substantially reduce the memory requirements. Median-beat tokenization requires only 0.08--0.16~GB and has an end-to-end latency of 3.87--4.64~ms per ECG, while HeartLang requires 0.22--0.66~GB and 4.59--5.22~ms per ECG. Their tokenizer-only costs are already 3.81 and 4.37~ms per ECG, respectively, indicating that heartbeat extraction dominates their wall-clock latency. 
This cost accompanies strong predictive performance. ECG-Byte is the slowest tokenizer at 10.36~ms per ECG, making its end-to-end latency approximately 11--12~ms across all backbones. BIOT-inspired tokenization is comparatively efficient, while the median beat provides a favourable predictive performance.

\begin{table}[h]
	\caption{Tokenizer-only efficiency on an NVIDIA A40 at batch size 32. Latency is the mean time per ECG over ten repetitions after three warm-up iterations.}
	\label{tab:tokenizer_efficiency}
	\centering
	\scriptsize
	\setlength{\tabcolsep}{25pt}
	\begin{tabular}{@{}lrr@{}}
		\toprule[1pt]
		Tokenizer & $L$ & Tokenization (ms/ECG) \\
		\midrule[0.6pt]
		Point-wise & 1000 & $<0.01$ \\
		Lead-wise & 12 & $<0.01$ \\
		Patching & 1500 & $<0.01$ \\
		Stochastic & 192 & 4.07 \\
		BIOT-inspired & 228 & 0.01 \\
		ECG-Byte & 1020 & 10.36 \\
		HeartLang & 256 & 4.37 \\
		Median beat & 60 & 3.81 \\
		\bottomrule[1pt]
	\end{tabular}
\end{table}

\section{Conclusion}
We found that tokenization is an important determinant of the ECG Transformer performance. Across four backbones, median-beat and HeartLang representations improve pooled mean macro-AUC from 0.823 to 0.891 relative to conventional point-wise and patch-wise inputs, while reducing mean peak training VRAM from 5.21 to 0.27~GB. The median beat offers the best overall performance--memory trade-off, whereas HeartLang with Reformer gives the best individual Transformer result. These findings show the value of preserving physiologically coherent heartbeat morphology for ECG tokenizers.

\section*{Acknowledgments}  
%
Jiawei Li and Antônio Horta Ribeiro are financially supported by the eSSENCE and SciLifeLab, with the
project "Digital Biomarkers from the Electrocardiogram using Artificial Intelligence";
and, by the Wallenberg AI, Autonomous Systems and Software Program (WASP) funded by Knut and Alice Wallenberg Foundation. 
This research was partially supported by \emph{Kjell och M{\"a}rta Beijer Foundation}.
This project has received funding from the European Research Council (ERC) under the European Union's Horizon Europe research and innovation programme through grant agreement no. 101054643.
Computations were enabled by resources provided by the National Academic Infrastructure for Supercomputing in Sweden (NAISS), partially funded by the Swedish Research Council through grant agreement no. 2022-06725.


\begin{correspondence}
Antonio H. Ribeiro\\
Box 337, 751 05, Uppsala, Sweden\\
antonio.horta.ribeiro@it.uu.se
\end{correspondence}

\end{document}